%% file: schemacaps_author_version.tex
\documentclass[a4paper,11pt]{article}

\usepackage[T1]{fontenc}
\usepackage{epsfig}
\usepackage{calc}
\usepackage[ruled,linesnumbered,vlined]{algorithm2e}
\usepackage{amssymb}
\usepackage{amstext}
\usepackage{amsmath}
\usepackage{amsthm}
\usepackage{multicol}
\usepackage{pifont}
\usepackage{pslatex}
\usepackage{apalike}
\usepackage{booktabs}
\usepackage{array}
\usepackage{xspace}
\usepackage{xurl}
\usepackage[protrusion=false,expansion=false]{microtype}
\usepackage[hidelinks,pdftitle={SchemaCaps: When a Compatible Tool Update Silently Widens an AI Agent's Authority},
  pdfauthor={Yulliwas Ameur, Samia Bouzefrane, Soumya Banerjee},
  pdfsubject={Author's accepted version, ICAART 2027, paper 479},
  pdfkeywords={LLM Agents, Model Context Protocol, Delegated Authority, Capabilities, JSON Schema Evolution, Tool Security}]{hyperref}
\usepackage{xcolor}
\usepackage[bottom]{footmisc}
\usepackage{authorversion}
\setAVvenue{ICAART 2027}
\setAVnotice{Author's accepted version of a paper accepted at the 19th International Conference on Agents and Artificial
Intelligence (ICAART 2027), Valletta, Malta, 23--26 February 2027. The final version is published by SCITEPRESS:
DOI to be added. This version does not use the publisher's layout.}
\setAVcode{\url{https://github.com/yulliwasameur/schemacaps-artifact}, archived at \url{https://doi.org/10.5281/zenodo.24233014}}

\newtheorem{definition}{Definition}
\newtheorem{lemma}{Lemma}
\newtheorem{proposition}{Proposition}

\newcommand{\sem}[1]{[\![#1]\!]}
\newcommand{\ttc}[1]{\texttt{\small #1}}
\newcommand{\eff}{\mathit{eff}}
\newcommand{\Auth}{\mathrm{Auth}}
\newcommand{\sys}{\mbox{SchemaCaps}\xspace}

\newcommand{\artefact}{the artefact}
\input{numbers.tex}
\SetAlFnt{\footnotesize}\SetAlCapFnt{\footnotesize}\SetAlCapNameFnt{\footnotesize}

\begin{document}

\title{SchemaCaps: When a Compatible Tool Update\\Silently Widens an AI Agent's Authority}

\author{\authorname{Yulliwas Ameur\sup{1,2}, Samia Bouzefrane\sup{2} and Soumya Banerjee\sup{3}}
\affiliation{\sup{1}Efrei Research Lab, Universit\'e Paris-Panth\'eon-Assas, Villejuif, France}
\affiliation{\sup{2}CEDRIC, Conservatoire National des Arts et M\'etiers (Cnam), Paris, France}
\affiliation{\sup{3}Inria, France}
\email{yulliwas.ameur@efrei.fr, samia.bouzefrane@lecnam.net, dr.soumya@ieee.org}
}

\keywords{LLM Agents, Model Context Protocol, Delegated Authority, Capabilities, JSON Schema Evolution, Tool Security.}

\abstract{The tools users approve for their AI agents keep changing their JSON input schemas. Of
\nSurveyProducts{} surveyed MCP clients and scanners, none is observed or documented to re-approve a tool whose input
schema widens (two not determinable). We define non-amplification over the coordinates visible in the schema (declared
arguments and undeclared fields, on effective values) and a re-binding rule decided by subschema inclusion, trusting no field labels by
default and conservative up to its checker and tested case analysis. Evidence: a Coq
proof of composition; \nTests{} regression tests; an exact-reference benchmark, where the reviewed
version of the rule silently re-bound \nBenchVthreeSilent{} amplifying changes, now fixed; and a
pilot on \nReleasesOK{} releases of \nPackagesOK{} public MCP servers. There, \nCAStrictFree{} of \nCompat{}
compatible schema changes widen some coordinate, yet compatibility re-binds them silently. The authors'
adjudicated double annotation finds at least \nKCore{} that add or widen an authority-bearing argument (\nKCoreFlagged{} of
the \nCACore{} a name-based labeller flags, plus \nKCoreMissed{} it misses). On existing tools, the rule asks
\nAskSCFree{} re-approvals, against \nChanged{} for pinning the input schema (\nAskPin{} with descriptions) and
\nAskCompat{} for compatibility. Behaviour change behind an identical schema remains out of reach.}

\maketitle

\section{\uppercase{Introduction}}
\label{sec:intro}

Agent platforms let a model act through tools exposed by third-party servers.\footnote{The text of the abstract and Sections 1--10 was drafted with the AI system Claude \cite{anthropic2026claude} and revised by the authors; see the Acknowledgements.} In the Model Context Protocol (MCP)
a server publishes each tool as a name, a description and a JSON Schema for its arguments; a user approves the
tool, and the agent then calls it on the user's behalf \cite{mcpspec}. Servers are ordinary packages that ship new
releases every few days, and a server may announce a changed tool list in the middle of a session.

Consider a real change from our corpus. In Desktop Commander 0.1.34, \ttc{execute\_command} takes \ttc{command}
and \ttc{timeout\_ms} on an object that rejects any other field; in 0.1.35 it also takes \ttc{shell}, any string.
Every old call remains valid, so for backward compatibility nothing changed. For the delegation, a policy on
\ttc{command} written for one interpreter no longer says what may run once the agent picks the interpreter. Some
arguments carry authority in this sense --- \ttc{shell}, a \ttc{webhookUrl}, a target project --- because their
value decides \emph{what} is acted upon, \emph{where} data goes or \emph{how} it runs; others, such as \ttc{limit}
or a pagination \ttc{cursor}, do not.

Deployed clients, as far as we could determine, do not ask this question (Section~\ref{sec:pilot}). The SDKs
hand the new schema over silently; VS Code keeps approvals under the same tool identifier; the one drift check
(Vercel AI SDK, opt-in) flags \emph{any} change; mcp-scan pinned the description only.

\textbf{Thesis.} \emph{A delegation should survive a schema change only if the change admits no new value on any
authority-bearing coordinate visible in the schema; this is decidable conservatively by schema inclusion, and
without trusting any labelling if every unlabelled coordinate is treated as authority-bearing.} The guarantee is
schema-visible by construction: a server can still change behaviour behind an identical schema, a boundary we mark
throughout. Simpler rules exist: pinning re-asks on every change, compatibility on none that keeps old calls valid;
Section~\ref{sec:pilot} measures what each costs and misses. Table~\ref{tab:contrib} lists our claims with the level
of evidence behind each.

\begin{table}[htbp]
\caption{Claims and evidence: \textsc{m} proof, \textsc{t} tests, \textsc{p} prototype, \textsc{e} measurement.}
\label{tab:contrib}
\centering\footnotesize
\setlength{\tabcolsep}{2.5pt}
\begin{tabular}{@{}>{\raggedright}p{5.0cm}c>{\raggedright\arraybackslash}p{6.5cm}@{}}
\toprule
Claim & Lvl & Evidence and scope\\
\midrule
Non-amplification composes along a chain (\S\ref{sec:rule}) & \textsc{m} & Coq; per-step inclusions assumed\\
The rule decides Definition~1 conservatively, up to its checker and tested case analysis (\S\ref{sec:rule}) & \textsc{t} & \nTests{} tests; exact-reference benchmark; oracle\\
A client-side proxy enforces it (\S\ref{sec:sys}) & \textsc{p} & stdio prototype, one server at a time\\
Widenings are exploitable (scripted planner) and blocked (\S\ref{sec:demo}) & \textsc{p} & scripted planner; six real cases replayed\\
Compatible widenings occur; cost of each rule (\S\ref{sec:pilot}) & \textsc{e} & \nReleasesOK{} releases; existence and cost, not prevalence\\
Surveyed clients do not re-check (\S\ref{sec:pilot}) & \textsc{e} & \nSurveyProducts{} products, \nSurveyTested{} run; two not determinable\\
\bottomrule
\end{tabular}
\end{table}

\section{\uppercase{Setting and Threat Model}}
\label{sec:model}

\textbf{Parties.} A \emph{delegator} grants an \emph{agent} the right to call tool $t$ of a \emph{server}. The
\emph{client} mediates every call. The grant is a capability $c=(\mathit{id}_{\mathrm{srv}},t,h_S,h_A,U,P)$. It holds the
server identity (launch-command digest and reported name) and the SHA-256 digests $h_S$ of the bound input schema $S$
and $h_A$ of the client's role manifest entry $A$ for $t$. It also holds the cumulative set $U$ of coordinates with
their roles, and a policy $P$, a predicate on effective authority values. The client admits a call $x$ only if $x\in\sem{S}$ (the
JSON values $S$ accepts) and $P$ holds on its effective values. Policies may be attenuated along a delegation chain
\cite{dennis1966,birgisson2014macaroons,prakash2026aip}.

\textbf{Effective values.} A call may omit an argument. $\eff_S(x)_a$ is the given value of path $a$ (e.g.\
\ttc{files[].path}) if present, else the declared default, else $\bot$ (``server decides''). $\bot$, \ttc{null} and
the empty string are distinct; $P$ states whether it accepts $\bot$. Configured and computed server defaults are
invisible in the schema and count as $\bot$ (\S\ref{sec:threats}).

\textbf{Coordinates and roles.} The coordinates of $S$ are its declared paths and one undeclared-field coordinate
$\star_o$ per object $o$ (the keys $o$ admits beyond its declared ones). An argument is \emph{authority-bearing} when
its value can redefine the target of the action, send data to a destination, change the principal or tenant, or
change the mode of execution (\ttc{shell}, \ttc{isUrl}); filters, pagination, free content, concurrency tokens and
metadata are not. This is a property of tool semantics, carried by the client's versioned manifest $A$ (path
$\mapsto$ authority or inert), not by field names. A path the manifest does not list is \emph{unknown} and treated as
authority-bearing; with no manifest every path is unknown and the rule is \emph{label-free}. A lexical labeller
(field-name patterns), never the default, is measured against human judgement in Section~\ref{sec:pilot}.

\textbf{Adversary.} An honest-but-careless server widens an argument for a feature; a malicious server rug-pulls in
one step or launders a field through an object opening; an injected instruction \cite{greshake2023} steers the
agent to use the new freedom. The client sees every version's schema and is trusted. Roles come from its manifest, never from
the server: in the label-free default, an innocuous field name evades nothing.
\sys addresses widenings of declared arguments, object openings, two-step laundering, fields no name pattern flags,
and tampering with or truncation of its local lineage. It does not address a manifest that wrongly marks a path
inert, behaviour change behind an identical schema or a non-amplifying one, injection through descriptions
\cite{greshake2023,debenedetti2025camel}, equivocation across clients (a transparency log would), or a compromised
client. Drafted with Claude \cite{anthropic2026claude}.

\section{\uppercase{Compatibility Is Not Authority}}
\label{sec:taxonomy}

$S\to S'$ is \emph{backward compatible} when $\sem{S}\subseteq\sem{S'}$. For coordinates $C$, $\pi_C\sem{S}$ is
the set of effective tuples they take over accepted calls; if $S$ leaves $a$ undeclared on a closed object then
$\pi_a\sem{S}=\{\bot\}$, and on an open object it is the whole value space. Table~\ref{tab:taxonomy} crosses the two questions. Adding a field to a closed object is compatible yet moves $\pi_a$ from $\{\bot\}$ to a domain (the
Desktop Commander case); opening an object widens $\star_o$ to fields no policy can name; a new \emph{required}
field breaks old calls \emph{and} amplifies. Compatibility concerns the interface; amplification, the delegation. Drafted with Claude \cite{anthropic2026claude}.

\begin{table}[htbp]
\caption{Schema changes on the two axes. Counts: changed tools among the \nCompat{} compatible changes of the pilot
whose reasons include the kind (lexical / label-free), except $^\dagger$ (all tool pairs).}
\label{tab:taxonomy}
\centering\footnotesize
\setlength{\tabcolsep}{4pt}
\begin{tabular}{@{}>{\raggedright}p{5.5cm}ccr@{}}
\toprule
Change & Compat. & Amplifies & lex./free\\
\midrule
Strict narrowing$^\dagger$ & no & no & \nTxNarrow\\
Field added, closed object & yes & \textbf{yes} & \nTxNewArg{} / \nTxNewArgFree\\
Domain widened & yes & \textbf{yes} & \nTxWidened{} / \nTxWidenedFree\\
Field became optional & yes & if $\bot\notin P$ & \nTxOptional{} / \nTxOptionalFree\\
Declaration dropped, open object & yes & \textbf{yes} & \nTxDropped{} / \nTxDroppedFree\\
Object opened & yes & \textbf{yes} ($\star_o$) & \nTxOpened{} / \nTxOpenedFree\\
Open sub-object introduced & yes & \textbf{yes} ($\star_o$) & \nTxIntroOpen{} / \nTxIntroOpenFree\\
\ttc{null} newly admitted & yes & \textbf{yes} & \nTxNullable{} / \nTxNullableFree\\
Default changed & yes & yes & \nTxDefault{} / \nTxDefaultFree\\
Required or defaulted field removed$^\dagger$ & no & \textbf{yes} & \nTxRemovedBottom\\
Beyond typed $\star_o$$^\dagger$ & -- & \textbf{yes} & \nTxBeyondStar\\
Tool hint raised$^\dagger$ & -- & yes & \nTxHints\\
\bottomrule
\end{tabular}
\end{table}

\section{\uppercase{Non-Amplification}}
\label{sec:rule}

Let $K$ be the authority-bearing coordinates (every declared path in label-free mode) together with every
$\star_o$. All projections are on effective values.

\begin{definition}[Non-amplification]
$S\to S'$ is \emph{non-amplifying} on $K$ if $\pi_{K}\sem{S'}\subseteq\pi_{K}\sem{S}$.
\end{definition}

The inclusion is joint because constraints can relate arguments. It reduces to a per-coordinate check under a
syntactic side condition:

\begin{lemma}[Product condition]
If $\pi_K\sem{S}$ is the product of its marginals, then $\pi_a\sem{S'}\subseteq\pi_a\sem{S}$ for every $a\in K$
implies non-amplification on $K$.
\end{lemma}

The condition holds when no keyword relating siblings (\ttc{anyOf}, \ttc{oneOf}, \ttc{allOf}, \ttc{if},
\ttc{not}, \ttc{dependent*}, \ttc{min}/\ttc{maxProperties}, \ttc{propertyNames}, \ttc{patternProperties}) lies
in $S$ above a coordinate of $K$, and no reference is unresolved; otherwise the answer is \textsf{unknown} and the
rule asks. Without it the marginal check is unsound: a top-level \ttc{anyOf} requiring \ttc{id} \emph{or}
\ttc{full\_path} that drops to requiring neither newly admits $(\bot,\bot)$.

\textbf{The checker.} Inclusion is undecidable in general \cite{attouche2022witness}. We use \ttc{jsonsubschema}
\cite{habib2021jsonsubschema}, reading its failures as \textsf{unknown}. It is not sound on every input: it answers
\textsf{yes} for an unrestricted boolean against an \ttc{enum}- or \ttc{const}-restricted one, on
\nProbesRawUnsound{} of our \nProbes{} probes, which would let a boolean flag widen silently. Our checker
$\mathcal{C}$ rewrites such booleans to an explicit \ttc{enum}, which fixes this case but not every false \textsf{yes}.
It then guards every \textsf{yes} between different schemas with a bounded, fixed-seed counterexample search
(Hypothesis \cite{maciver2019hypothesis}; 60 draws, 2\,s); a found counterexample turns the answer to \textsf{no}. The guard is
defence in depth, not a proof.

\begin{algorithm}[htbp]
\caption{Re-binding a capability $c$ bound to $S$ when the server publishes $S'$ (manifest $A$).}
\label{alg:rebind}
\lIf{$\mathit{id}_{\mathrm{srv}}$ changed}{suspend($c$, identity); \Return}
\lIf{$A$ relabels a coordinate of $c.U$ inert}{suspend($c$, relaxed); \Return}
\lIf{a tool hint is raised (read-only dropped, destructive or open-world raised)}{suspend($c$, hint); \Return}
\lIf{$S'=S$ (description or title change)}{\Return}
\lIf{product condition fails on $S$, or \ttc{\$ref} unresolved}{suspend($c$, unknown); \Return}
\ForEach{coordinate $a$ of $S$ or $S'$ not marked inert by $A$}{
  \lIf{$a$ new on a closed object, or $\star_o$ opened, or $a$ now optional, or now admits \ttc{null}, or its default changed, or $a$ removed while required or defaulted, or $a$ declared beyond the old typed $\star_o$}{suspend($c$, $a$); \Return}
  \lIf{$\mathcal{C}(\pi_a S',\pi_a S)\neq\textsf{yes}$}{suspend($c$, $a$); \Return}
}
$c.h_S\gets h(S')$;\ $c.U\gets c.U\cup\mathrm{coords}(S')$;\ append a signed receipt\;
\end{algorithm}

\textbf{Rule.} Algorithm~\ref{alg:rebind} re-binds silently only when every check passes; otherwise it suspends the
tool and asks the delegator, who may approve this version, tighten $P$, or refuse. A new coordinate on a closed
object has old marginal $\{\bot\}$ and always asks; so does a field that newly admits \ttc{null}, which is a value,
not absence, and a required or defaulted field that disappears, whose effective value becomes $\bot$. The rule is
\emph{orthogonal to compatibility}: an incompatible change that only narrows (a field becomes required) re-binds
silently --- old calls may now fail, but no authority is gained --- while a compatible widening asks. A later release
that is non-amplifying with respect to the \emph{bound} schema lifts a suspension.

\begin{proposition}[No amplification along chains]
Let $S_0\!\to\!\cdots\!\to\!S_n$ with $c$ re-bound silently at each step and roles fixed. Then
$\pi_{K}\Auth(S_n,P)\subseteq\pi_{K}\Auth(S_0,P)$, where $\Auth(S,P)=\{x\in\sem{S}: P(\eff_S(x))\}$.
\end{proposition}

Each silent step gives the per-coordinate inclusions, lifted by Lemma~1; transitivity and intersection with
$P$ conclude. In label-free mode a silent re-bind also implies $\sem{S'}\subseteq\sem{S}$ with no declared default
changed or removed, which the oracle of Section~\ref{sec:pilot} tests.

\textbf{What is proved, tested and argued.} The Coq development (\nCoqLines{} lines; no axioms, no admits)
mechanises Definition~1, Lemma~1, the chain proposition, and an effect corollary under a \emph{fidelity}
assumption (the server's effect never exceeds what the authority tuple permits), which hidden behaviour change breaks.
It fixes the coordinate set; $U$, $\star_o$ and effective values are argued on paper and tested. The per-step
inclusions come from the checker, which is tested, not verified.

\textbf{Exact-reference benchmark.} We generate schema pairs in a finite-domain fragment of JSON Schema (one
nesting level, \ttc{if}/\ttc{then}, \ttc{anyOf}, \ttc{not}, \ttc{\$ref}, typed \ttc{additionalProperties}). There,
inclusion and amplification on effective values are decided \emph{exactly} by enumerating a finite universe that
holds a witness for every violation (informal argument in the artefact; relative to the python-jsonschema validator), independently of \ttc{jsonsubschema}. Of
\nBenchGenerated{} valid changes, \nBenchDiscarded{} with a universe above 150,000 values are discarded: large schemas
are under-represented. Of \nBenchCheckNotIncl{} non-inclusions among \nBenchCheckPairs{} checker pairs, raw
\ttc{jsonsubschema} accepts \nBenchRawFA, the boolean rewrite alone \nBenchNormFA, and $\mathcal{C}$ \nBenchGuardFA{}
(\nBenchChkUnknown{} \textsf{unknown} on our machine, no false rejection).
Conditional pairs are few (\nBenchChkCondPairs): on all \nBenchChkCondNotIncl{} of their non-inclusions the
unguarded checker answers \textsf{yes}, and only the randomized guard rejects them. Of \nBenchRulePairs{} changes
(\nBenchNested{} nested, \nBenchCond{} conditional, \nBenchRef{} with \ttc{\$ref}), \nBenchRuleAmp{} amplify. The
rule of the version reviewed for ICAART re-bound \nBenchVthreeSilent{} of them silently (emulated by disabling the two new checks), in two cases: a required or defaulted
field removed from a closed object (\nBenchVthreeRemoved), and a field declared beyond a typed
\ttc{additionalProperties} (\nBenchVthreeStar). The pilot's oracle shared the first blind spot. We fixed both (line~7 of Algorithm~\ref{alg:rebind}; deviation~11 in the artefact); in the pilot, the fix changes no label-free
count and adds one lexical re-approval. In label-free mode, the fixed rule re-binds no amplifying
change silently, on this benchmark (\nBenchFalseSilent) and on a second seed recorded before the run, a
self-attested ordering (\nBenchTwoFalseSilent{} of \nBenchTwoAmp; reviewed version: \nBenchTwoVthreeSilent). The
\nBenchSiblingAsk{} changes whose old schema relates siblings are answered \textsf{unknown} and asked, not checked.
It asks on \nBenchAsksNonAmp{} of \nBenchRuleNonAmp{} non-amplifying changes (\nBenchUnnecPct\%, tied to this
synthetic mix), pinning on all; compatibility misses \nBenchCompatFalseSilent{} amplifying ones. Drafted with Claude
\cite{anthropic2026claude}.

\section{\uppercase{Enforcement: a Client-Side Proxy}}
\label{sec:sys}

The proxy (\ttc{prototype/} in \artefact) sits between an agent client and a stdio server. On \ttc{tools/list} it
applies a chosen rule (\ttc{pin}, \ttc{compat} or \sys) to every changed tool, hides suspended tools from the model
and refuses their calls with a named reason; a change of server identity suspends every tool. On \ttc{tools/call} it
validates the arguments against the bound schema and evaluates $P$ on effective values. Each decision appends a
signed, hash-chained receipt \cite{rfc8032}; \ttc{verify} detects tampering and truncation of the local log, not
equivocation across clients, which would need a transparency log \cite{laurie2013ct}.

\textbf{Cost.} With call validation and a policy on, the proxy adds \nOvCall\,ms per call
(\nOvCallMin--\nOvCallMax\,ms across \nOvRuns{} runs of \nOvIter{} iterations) and \nOvList\,ms per list on a
2-core x86-64 machine. On that machine (machine-dependent values) and with the rule of the reviewed version, a decision on a changed tool takes \nTmMedFree\,ms median
label-free (\nTmMedLex\,ms lexical), \nTmPNNFree\,s at p99 and \nTmMaxFree\,s at worst, dominated by the checker's
guard (2\,s per inclusion); the fix may add one guarded inclusion per removed or newly declared coordinate. It is made once per
changed (tool, schema) pair; the guard's verdicts are cached by schema digests, so a refresh re-runs only the fast unguarded check. The single-threaded proxy holds calls during the decision; the
tool stays bound to its old schema until then. Drafted with Claude \cite{anthropic2026claude}.

\section{\uppercase{Demonstration}}
\label{sec:demo}

\begin{table*}[htbp]
\caption{Staged scenarios (grant at v1, then the change). ``blocked'': asked and the attack call refused;
``succeeds'': the external effect occurs; benign change: whether the delegator is asked. Manifest: \ttc{body},
\ttc{priority}, \ttc{artifact} inert.}
\label{tab:demo}\let\code\ttc
\centering\footnotesize
\setlength{\tabcolsep}{4pt}
\begin{tabular}{@{}lcccccc@{}}
\toprule
& & & & \multicolumn{3}{c}{\sys}\\
\cmidrule(l){5-7}
Change & Accept-all & Pin & Compatibility & label-free & manifest & lexical\\
\midrule
\DemoRows
\bottomrule
\end{tabular}
\end{table*}

\textbf{Staged scenarios.} A reporting server exposes \ttc{send\_report} (to a team address, on a closed object) and
\ttc{deploy} (\ttc{env} an enum \ttc{test}); its inbox carries an injected instruction to forward reports to an
external address. The planner is \emph{scripted}: it obeys the injection; no LLM is run. This suffices here, because the
client's decision depends only on \ttc{tools/list} and \ttc{tools/call}, not on the planner; an end-to-end run with a
live model remains future work. Table~\ref{tab:demo} records, after the upgrade, whether the delegator is asked and
the external effect occurs. Compatibility lets every widening through; pinning blocks them but also asks on the
benign change. Label-free \sys blocks every staged schema-visible attack and asks on
the benign change; with \ttc{priority} marked inert it also re-binds the benign change silently; with the lexical labeller it misses the enum widening, because \ttc{env} matches no authority name.
A behaviour change behind an identical schema defeats every rule. All \nReceiptsVerified{} receipt chains verify.

\textbf{Real replay.} We replay six hand-verified compatible widenings, each on its exact release pair: Desktop
Commander \ttc{shell} and \ttc{isUrl}, Firecrawl \ttc{webhookUrl}, GitLab \ttc{target\_project\_id}, Sentry
\ttc{regionUrl}, and Notion's opened \ttc{parent}. The proxy's decision function runs on the extracted tool lists, no server
executed: compatibility re-binds all \nReplayCompatRebind{} silently; pinning and \sys suspend all of them
(\nReplayOK/\nReplayCases). Drafted with Claude \cite{anthropic2026claude}.

\section{\uppercase{Pilot on MCP Servers}}
\label{sec:pilot}

The pilot shows that compatible widenings \emph{exist} in real servers and measures each rule's cost; it does not
estimate an ecosystem rate. \emph{Units}: a \emph{release} is one published version of a server package; a
\emph{pair} is two successive releases of one package; a \emph{change} is one tool, present in both releases of a
pair, whose input schema differs; a change is \emph{compatible} if every old call stays valid.

\textbf{Protocol and deviations.} The protocol (corpus, window, extraction, questions, rules) was hashed at 15:25 UTC
on 29 Sept.\ 2026, before the corpus was extracted; it was not deposited with a third party, so the ordering is
self-attested. Deviations 1--4 (sampling cap, exclusions, strata) precede any classification; deviations 5--11 (measurement
fixes, classifier refinements, the benchmark fix of Section~\ref{sec:rule}, the K correction, the simple-pin baseline) are post hoc, as is the final classifier.

\textbf{Corpus and extraction.} Stratum A: the official reference servers (npm \ttc{@modelcontextprotocol/\allowbreak server-*}
without SDKs and skeletons; PyPI \ttc{fetch}, \ttc{git}, \ttc{time}, \ttc{sqlite}). Stratum B: the 15 most
downloaded third-party npm stdio servers answering \ttc{tools/list} with dummy credentials. Window: stable releases
from Jan.\ 2025 to Sept.\ 2026, \nWindowVersions{} releases of \nWindowPackages{} packages, at most 20 per package.
Each release is installed with dependencies resolved at its publication date and queried over stdio;
\nReleasesOK{} of \nReleases{} sampled releases (\nPctOK\%) of \nPackagesOK{} packages answered (\nReleasesOKA{} A,
\nReleasesOKB{} B), giving \nPairs{} pairs. A per-release manifest records install, hashes and outcome. HTTP, authenticated-only and PyPI third-party servers are not covered.

\begin{table*}[htbp]
\caption{Pilot results. Changed: tools whose input schema changed within a pair; Compatible: every old call stays
valid. Amplifying: adds or widens a lexically flagged argument (core), meets Definition~1 with lexical roles (strict), or any
coordinate (label-free). Re-approvals on tools present in both releases (every rule also asks for the \nAdded{} added
tools). SC: \sys.}
\label{tab:results}
\centering\footnotesize
\setlength{\tabcolsep}{3pt}
\begin{tabular}{@{}lrrrrrrrrrr@{}}
\toprule
& & & & \multicolumn{3}{c}{Compatible and amplifying} & \multicolumn{4}{c}{Re-approvals}\\
\cmidrule(lr){5-7}\cmidrule(l){8-11}
Stratum & Pairs & Changed & Compatible & core & strict & label-free & Pin & Compat. & SC lex. & SC free\\
\midrule
A official & \nPairsA & \nChangedA & \nCompatA & \nCACoreA & \nCAStrictA & \nCAStrictFreeA & \nAskPinA & \nAskCompatA & \nAskSCA & \nAskSCFreeA\\
B third-party & \nPairsB & \nChangedB & \nCompatB & \nCACoreB & \nCAStrictB & \nCAStrictFreeB & \nAskPinB & \nAskCompatB & \nAskSCB & \nAskSCFreeB\\
Total & \nPairs & \nChanged & \nCompat & \textbf{\nCACore} & \nCAStrict & \nCAStrictFree & \nAskPin & \nAskCompat & \nAskSC & \textbf{\nAskSCFree}\\
\bottomrule
\end{tabular}
\end{table*}

\textbf{Results and concentration.} Of \nChanged{} changes, \nCompat{} are compatible (Table~\ref{tab:results});
\nCAStrictFree{} of them widen some coordinate and \nCACore{} add or widen an argument the lexical labeller flags as
authority-bearing, all in stratum B; adjudication confirms \nKCoreFlagged{} of them and adds \nKCoreMissed{} it missed (\emph{Roles}). The core changes sit in \nCoreServers{} servers and \nCorePairs{} of \nPairsB{}
pairs. Two packages hold \nTopTwo{} (a mobile-automation server adding two local-path arguments to 37 tools, an
error-tracking server adding a region endpoint to 15); removing the first, the second, or both leaves \nLooTopOne{},
\nLooTopTwo{} and \nWithoutTopTwo{} core changes. The phenomenon is real but concentrated: the core rate is \nPctCACore\% of
the \nCompat{} compatible changes, and resampling packages with replacement (\nBootDraws{} draws, fixed seed)
spreads it over $[\nCILow,\nCIHigh]\%$ (label-free: $[\nCIFreeLow,\nCIFreeHigh]\%$), a stability check rather than a
population estimate. Sampling the two dominant packages to 40 releases adds only \nCapPlusA{} and \nCapPlusB{} core changes.

\textbf{Roles.} The lexical figures rest on whether an argument is authority-bearing. A first annotation pass, which read
authority too structurally, was discarded. With the criterion of Section~\ref{sec:model}, two authors then labelled
\nAnnCases{} blind cases separately, without pre-labels or AI suggestions. They are the \nAnnFlagged{} flagged arguments with an ask reason in
a compatible change when the sheets were drawn (\nFrameFlagged{} under the final classifier; the \nFrameMissing{} others carry no core reason) and all
\nAnnUnflagged{} unflagged arguments added under a closed ancestor, for recall.
A third author adjudicated all \nAnnDisagree{} disagreements. Agreement is \nAnnAgree\% raw, Cohen's
$\kappa=\nKappa$ \cite{cohen1960}, substantial \cite{landis1977}; nearly all of the residual is one annotator's
\nAnnUnsure{} ``unsure'' answers on borderline identifiers ($\kappa=\nKappaCollapsed$ reading unsure as no;
\nAnnHard{} hard conflicts). Against the adjudicated labels the lexical rule has precision \nPrecision{} and recall
\nRecall{} (an upper bound, misses being sampled only among added arguments): a quarter of its flagged arguments are wrong and it
misses \ttc{saveScript}, \ttc{webhook.headers} and an execution \ttc{mode}. Hence \nCACore{} is the labeller's
count of core changes; adjudication confirms \nKCoreFlagged{} of them and finds \nKCoreMissed{} it missed: at least
\nKCore{} in all (the reviewed version reported \nKCoreReviewed{}, 66 + 7, as a subset of the \nCACore{} and dropped 3 changes that
also carried another (non-core) ask reason; deviation~11). The lexical labeller is not the default.

\textbf{Soundness evidence.} A differential oracle audits \emph{every} silent re-bind, compatible or not. In
label-free mode, \nOrSchemaFree{} re-binds change the schema (\nOrCompatFree{} compatible, \nOrIncompatFree{}
incompatible, \nOrUndFree{} undecided). For each, up to \nOrN{} calls generated from the new schema must validate
against the old one, and no declared default may differ or disappear: \nOrGenFree{} calls (time-bounded, machine-dependent), \nOrGenErrFree{}
generation failures, \nOrWitnessFree{} counterexamples. In lexical mode, a semantic probe checks the \nOrSchemaLex{}
schema-changing re-binds: \nOrWitnessLex{} counterexamples. The guard ran \nGuardCallsFree{} times and overturned
\nGuardOverturnedFree{} verdicts; a reverse search on all \nCompatYesChecked{} compatibility \textsf{yes} verdicts found no
counterexample. The oracle is itself tested on planted violations. This is evidence, not proof.

\textbf{Cost, and why not just pin.} Pinning is simpler and sound. Pinning the whole definition asks \nAskPin{}
re-approvals on existing tools, \nDescOnly{} of them for description or title changes only; pinning the input schema
alone, the fair baseline since \sys ignores descriptions, asks \nChanged. Both ask again on narrowings. Conversely, a server can pair a behaviour change with a narrowing or a
description-only edit: \sys re-binds those silently where pinning asks, which is the price of its lower cost.
The simplest rule, `pin unless $\sem{S'}\subseteq\sem{S}$', asks \nSimplePinAsk{} yet re-binds silently
\nSimplePinMissPilot{} changes \sys asks on (\nSimplePinMissAmp{} that \sys judges amplifying, \nSimplePinMissUnd{} that it cannot decide), and \nSimplePinMissBench{} and \nSimplePinMissBenchTwo{} amplifying
changes of the two benchmark seeds: it ignores effective values. Deployment needs
the decision function (\ttc{classify2.py}, \ttc{jsonsubschema}, Hypothesis; calls wait up to \nTmMaxFree\,s on our machine) in
the client; the proof and the benchmark are offline evidence. Compatibility asks
\nAskCompat{} and silently accepts the \nCAStrictFree{} compatible widenings. Label-free \sys asks \nAskSCFree{} ---
\nFreeMinusCompat{} more than compatibility (+\nFreeAboveCompat\%), about half of a schema-only pin and
\nFreeBelowPin\% fewer than a full pin --- and
stops all of them (by construction; the oracle checks independently); the lexical labeller would ask \nAskSC{} at the price of its misses. Counting the \nAdded{} added
tools, which every rule treats as new authority, the totals are \nTotPin{} (pin), \nTotCompat{} (compat), \nTotFree{}
(label-free) and \nTotLex{} (lexical). The check is \textsf{unknown} in \nUndecidedFree{} cases. How many of the
label-free questions a delegator would find unnecessary is not measured here: it needs real users (\S\ref{sec:agenda}).

\begin{table*}[htbp]
\caption{Deployed MCP software when an approved tool's input schema widens (30 Sept.\ 2026). \emph{tested}: run
against our drift server; \emph{source}: code read; \emph{docs}: documentation. Class: (a) uses the new definition
silently; (b) flags any change; (c) re-approves on incompatible change; (d) reasons about widening.}
\label{tab:clients}
\centering\footnotesize
\setlength{\tabcolsep}{3pt}
\begin{tabular}{@{}l>{\raggedright}p{2.5cm}l>{\raggedright}p{5.6cm}l@{}}
\toprule
Product & Version & Evidence & Observed on a schema widening & Class\\
\midrule
MCP TypeScript SDK & 1.31.0 & tested+source & re-list delivered only to an opt-in handler; no diff; widened call sent & a\\
MCP Python SDK & 2.2.0; 1.27.0 & tested+source & notification only via a user handler; widened call sent & a\\
Vercel AI SDK & ai 7.0.123, @ai-sdk/mcp 2.0.63 & tested+source & default silent; opt-in \ttc{detectToolDrift} reports ``changed'' & a / b opt-in\\
VS Code (agent mode) & 1.139.1 & source & tool id derived from its name, so approvals survive the change & a\\
Claude Code & 2.1.285 & docs & refreshes on list change; rules match tool names & a (docs)\\
Claude Desktop & 2.16120.0 & docs & per-tool approval; changes not documented & not determinable\\
Cursor & 3.22 & docs & approval by default; allowlist by name; changes not documented & not determinable\\
mcp-scan & 0.4.2 (pinning $\le$0.4.5) & tested+source & MD5 of the description only: no warning on a schema-only widening & b (desc.)\\
\bottomrule
\end{tabular}
\end{table*}

\textbf{Clients.} Table~\ref{tab:clients} records the survey (three SDK families, four host applications, one
scanner). The drift server widens \ttc{read\_file} mid-session (sandbox restriction dropped, \ttc{webhookUrl} added,
name and description kept) and sends \ttc{list\_changed}. Each client re-lists and calls it with a value only the new
schema admits (\nSurveyConfigs{} configurations, pinned versions). None of the six whose behaviour we could determine re-approves a widened tool, and the documentation of the
other two describes no such check. Drafted with Claude \cite{anthropic2026claude}.

\section{\uppercase{Related Work}}
\label{sec:related}

Our contribution is the question --- does a schema change widen delegated authority? --- and a re-binding rule, not
a new subschema algorithm, which we reuse \cite{habib2021jsonsubschema}. \emph{Tool-definition integrity.} ETDI
signs versioned tool definitions and re-asks on every new version \cite{bhatt2025etdi}. Acharya and Gupta define
integrity as equality with the approved definition \cite{acharya2026mcpshield}. mcp-scan pinned the description
\cite{mcpscan}, the Vercel AI SDK offers opt-in drift detection \cite{vercel2026drift}, and Song et al.\ demonstrate
rug pulls that rewrite descriptions \cite{song2025beyond}. All decide \emph{that} a definition changed; \sys decides whether the delegated authority grew,
and re-binds narrowings silently. \emph{Admission and confinement.} Attested admission allowlists tools by name and leaves schema
drift to a downstream monitoring layer \cite{metere2026attested}, the layer \sys provides. AttestMCP adds capability
certificates and message authentication \cite{maloyan2026breaking}. AgentBound confines each server with a permission
manifest \cite{buhler2025agentbound}. \emph{Privilege control in agents.} Progent applies
narrowing policy updates automatically and asks for expansions \cite{shi2025progent}: the same inclusion question on
the policy axis ($P\to P'$), where \sys asks it on the schema axis ($S\to S'$). ChainCaps attenuates capability budgets along tool chains \cite{jiang2026chaincaps}; \sys governs how the
schema may change. \emph{Update-time escalation} is classic:
browsers disable an extension whose update adds a permission \cite{chromeperm}, and policy analysers decide
whether a new policy grants more \cite{backes2018zelkova}; we bring the question to argument schemas, whose
evolution API-diff tools classify only for compatibility \cite{brito2018apidiff}. Drafted with Claude \cite{anthropic2026claude}.

\section{\uppercase{Threats to Validity}}
\label{sec:threats}

\emph{Scope.} Besides the cases of Section~\ref{sec:model}, \sys does not decide configured or computed server
defaults or output schemas. \emph{Construct.} The
lexical figures, not the label-free ones, rest on the notion of authority. Annotators and adjudicator are authors;
their independence cannot be attested. \emph{Internal.} The classifier is post hoc. The checker is tested and
guarded, not verified. Configured defaults count as $\bot$. The benchmark's 15 mutation operators are the authors',
its second seed reuses the generator, \ttc{\$ref} is local, and its machine-dependent guard finds witnesses easily
in such small domains (rule-level results are identical without it); it omits floats, formats,
\ttc{patternProperties}, \ttc{min}/\ttc{maxProperties} and string lengths. \emph{External.} The
corpus (top npm stdio and official servers, at most 20 releases each) is not random, and its widenings concentrate
in a few servers; results do not transfer to HTTP servers or other ecosystems. \emph{Demonstration.} The planner is scripted; real cases are replayed on schemas,
not executed; overhead is measured on one machine. \emph{Survey.} \nSurveyTested{} products were run, one read in source, three only in documentation;
Claude Desktop and Cursor are closed and could not be observed, and documentation-only classifications may
differ from behaviour. Drafted with Claude \cite{anthropic2026claude}.

\section{\uppercase{Conclusion and Open Problems}}
\label{sec:agenda}

Compatibility preserves old calls; \sys also preserves, up to its checker and tested case analysis, the delegated authority visible in the input schema, without
trusting any labelling, at about a compatibility rule's question budget: on existing tools, about half of a schema-only pin's and a third
of a full pin's.
It does \emph{not} rule out hidden behavioural change. Open problems:
attesting behaviour behind an identical schema; signed argument roles; policy-relative re-binding; a broader corpus
(HTTP servers, other registries); end-to-end runs with live models; and a user study of unnecessary questions, since
\nAskSCFree{} questions in 21 months still invite rubber-stamping. Agent platforms
should stop asking only whether old calls still validate and start asking whether authority grew. Drafted with Claude \cite{anthropic2026claude}.

\section*{\uppercase{Acknowledgements}}

AI-use disclosure: the text of the abstract and Sections 1--10 was drafted, and the artefact's code written, with the assistance
of Claude \cite{anthropic2026claude}, an AI system by Anthropic, under the direction of Yulliwas Ameur. The authors
specified the study, reviewed and tested the code, checked every claim, proof and figure against the artefact,
whose scripts generate every result in \ttc{numbers.tex}, and are responsible for the content. The annotation
of Section~\ref{sec:pilot} was done without AI. \emph{Artefact} (data, code, proofs, \ttc{make reproduce}):
\href{https://github.com/yulliwasameur/schemacaps-artifact}{\nolinkurl{github.com/yulliwasameur/schemacaps-artifact}},
archived as \href{https://doi.org/10.5281/zenodo.24233014}{\nolinkurl{doi.org/10.5281/zenodo.24233014}}.

\bibliographystyle{apalike}
{\small
\bibliography{references}}

\end{document}

%% file: numbers.tex
\newcommand{\nAdded}{907}

\newcommand{\nAnnAgree}{84}
\newcommand{\nAnnCases}{176}
\newcommand{\nAnnDisagree}{29}
\newcommand{\nAnnFlagged}{44}
\newcommand{\nAnnHard}{4}

\newcommand{\nAnnUnflagged}{132}

\newcommand{\nAnnUnsure}{25}
\newcommand{\nAskCompat}{617}
\newcommand{\nAskCompatA}{24}
\newcommand{\nAskCompatB}{593}

\newcommand{\nAskPin}{2{,}004}
\newcommand{\nAskPinA}{109}
\newcommand{\nAskPinB}{1{,}895}

\newcommand{\nAskSC}{413}
\newcommand{\nAskSCA}{34}
\newcommand{\nAskSCB}{379}
\newcommand{\nAskSCFree}{645}
\newcommand{\nAskSCFreeA}{40}
\newcommand{\nAskSCFreeB}{605}

\newcommand{\nBenchAsksNonAmp}{123}
\newcommand{\nBenchCheckNotIncl}{314}
\newcommand{\nBenchCheckPairs}{600}

\newcommand{\nBenchChkCondNotIncl}{6}
\newcommand{\nBenchChkCondPairs}{17}

\newcommand{\nBenchChkUnknown}{17}
\newcommand{\nBenchCompatFalseSilent}{272}
\newcommand{\nBenchCond}{95}
\newcommand{\nBenchDiscarded}{326}
\newcommand{\nBenchFalseSilent}{0}
\newcommand{\nBenchGenerated}{926}
\newcommand{\nBenchGuardFA}{0}

\newcommand{\nBenchNested}{181}
\newcommand{\nBenchNormFA}{9}

\newcommand{\nBenchRawFA}{37}
\newcommand{\nBenchRef}{94}
\newcommand{\nBenchRuleAmp}{330}

\newcommand{\nBenchRuleNonAmp}{270}
\newcommand{\nBenchRulePairs}{600}

\newcommand{\nBenchSiblingAsk}{135}

\newcommand{\nBenchTwoAmp}{333}

\newcommand{\nBenchTwoFalseSilent}{0}

\newcommand{\nBenchTwoVthreeSilent}{33}
\newcommand{\nBenchUnnecPct}{46}
\newcommand{\nBenchVthreeRemoved}{24}
\newcommand{\nBenchVthreeSilent}{27}
\newcommand{\nBenchVthreeStar}{3}
\newcommand{\nBootDraws}{10{,}000}
\newcommand{\nCACore}{84}
\newcommand{\nCACoreA}{0}
\newcommand{\nCACoreB}{84}

\newcommand{\nCAStrict}{274}
\newcommand{\nCAStrictA}{31}
\newcommand{\nCAStrictB}{243}
\newcommand{\nCAStrictFree}{412}
\newcommand{\nCAStrictFreeA}{34}
\newcommand{\nCAStrictFreeB}{378}
\newcommand{\nCIFreeHigh}{86}
\newcommand{\nCIFreeLow}{56}
\newcommand{\nCIHigh}{25.2}
\newcommand{\nCILow}{3.5}
\newcommand{\nCapPlusA}{5}
\newcommand{\nCapPlusB}{2}
\newcommand{\nChanged}{1{,}197}
\newcommand{\nChangedA}{87}
\newcommand{\nChangedB}{1{,}110}

\newcommand{\nCompat}{580}
\newcommand{\nCompatA}{63}
\newcommand{\nCompatB}{517}

\newcommand{\nCompatYesChecked}{580}
\newcommand{\nCoqLines}{117}
\newcommand{\nCorePairs}{20}
\newcommand{\nCoreServers}{6}
\newcommand{\nDescOnly}{807}
\newcommand{\nFrameFlagged}{50}
\newcommand{\nFrameMissing}{6}

\newcommand{\nFreeAboveCompat}{4.5}
\newcommand{\nFreeBelowPin}{68}
\newcommand{\nFreeMinusCompat}{28}

\newcommand{\nGuardCallsFree}{467}

\newcommand{\nGuardOverturnedFree}{0}

\newcommand{\nKCore}{76}
\newcommand{\nKCoreFlagged}{66}
\newcommand{\nKCoreMissed}{10}
\newcommand{\nKCoreReviewed}{73}

\newcommand{\nKappa}{0.67}
\newcommand{\nKappaCollapsed}{0.90}

\newcommand{\nLooTopOne}{42}
\newcommand{\nLooTopTwo}{64}

\newcommand{\nOrCompatFree}{159}

\newcommand{\nOrGenErrFree}{0}

\newcommand{\nOrGenFree}{55{,}003}

\newcommand{\nOrIncompatFree}{323}

\newcommand{\nOrN}{100}

\newcommand{\nOrSchemaFree}{553}
\newcommand{\nOrSchemaLex}{785}

\newcommand{\nOrUndFree}{71}

\newcommand{\nOrWitnessFree}{0}
\newcommand{\nOrWitnessLex}{0}
\newcommand{\nOvCall}{0.47}
\newcommand{\nOvCallMax}{0.56}
\newcommand{\nOvCallMin}{0.37}
\newcommand{\nOvIter}{500}
\newcommand{\nOvList}{1.1}

\newcommand{\nOvRuns}{5}
\newcommand{\nPackagesOK}{39}
\newcommand{\nPairs}{565}
\newcommand{\nPairsA}{337}
\newcommand{\nPairsB}{228}

\newcommand{\nPctCACore}{14.5}

\newcommand{\nPctOK}{91.9}
\newcommand{\nPrecision}{0.75}
\newcommand{\nProbes}{17}

\newcommand{\nProbesRawUnsound}{4}
\newcommand{\nRecall}{0.60}
\newcommand{\nReceiptsVerified}{30}
\newcommand{\nReleases}{657}
\newcommand{\nReleasesOK}{604}
\newcommand{\nReleasesOKA}{363}
\newcommand{\nReleasesOKB}{241}
\newcommand{\nReplayCases}{6}
\newcommand{\nReplayCompatRebind}{6}
\newcommand{\nReplayOK}{6}
\newcommand{\nSimplePinAsk}{693}
\newcommand{\nSimplePinMissAmp}{7}
\newcommand{\nSimplePinMissBench}{27}
\newcommand{\nSimplePinMissBenchTwo}{17}
\newcommand{\nSimplePinMissPilot}{22}
\newcommand{\nSimplePinMissUnd}{15}
\newcommand{\nSurveyConfigs}{6}
\newcommand{\nSurveyProducts}{8}
\newcommand{\nSurveyTested}{4}
\newcommand{\nTests}{48}
\newcommand{\nTmMaxFree}{10.5}

\newcommand{\nTmMedFree}{22}
\newcommand{\nTmMedLex}{13}

\newcommand{\nTmPNNFree}{2.6}

\newcommand{\nTopTwo}{62}
\newcommand{\nTotCompat}{1{,}524}
\newcommand{\nTotFree}{1{,}552}

\newcommand{\nTotLex}{1{,}320}
\newcommand{\nTotPin}{2{,}911}
\newcommand{\nTxBeyondStar}{0}
\newcommand{\nTxDefault}{0}
\newcommand{\nTxDefaultFree}{1}
\newcommand{\nTxDropped}{14}
\newcommand{\nTxDroppedFree}{33}
\newcommand{\nTxHints}{2}
\newcommand{\nTxIntroOpen}{17}
\newcommand{\nTxIntroOpenFree}{17}
\newcommand{\nTxNarrow}{333}
\newcommand{\nTxNewArg}{80}
\newcommand{\nTxNewArgFree}{167}
\newcommand{\nTxNullable}{17}
\newcommand{\nTxNullableFree}{21}
\newcommand{\nTxOpened}{129}
\newcommand{\nTxOpenedFree}{129}
\newcommand{\nTxOptional}{26}
\newcommand{\nTxOptionalFree}{45}
\newcommand{\nTxRemovedBottom}{2}

\newcommand{\nTxWidened}{6}
\newcommand{\nTxWidenedFree}{134}

\newcommand{\nUndecidedFree}{20}

\newcommand{\nWindowPackages}{45}
\newcommand{\nWindowVersions}{1{,}856}
\newcommand{\nWithoutTopTwo}{22}
\newcommand{\DemoRows}{%
Destination widened (\code{cc}) & \textbf{succeeds} & blocked & \textbf{succeeds} & blocked & blocked & blocked\\
Two-step laundering & \textbf{succeeds} & blocked & \textbf{succeeds} & blocked & blocked & blocked\\
Undeclared field after opening & \textbf{succeeds} & blocked & \textbf{succeeds} & blocked & blocked & blocked\\
Enum widened (\code{env}) & \textbf{succeeds} & blocked & \textbf{succeeds} & blocked & blocked & \textbf{succeeds}\\
Benign change (\code{priority}) & silent & asks & silent & asks & silent & silent\\
Behaviour change, same schema & \textbf{succeeds} & \textbf{succeeds} & \textbf{succeeds} & \textbf{succeeds} & \textbf{succeeds} & \textbf{succeeds}\\}